\documentclass[conference]{IEEEtran}
\usepackage{times}
\usepackage[T1]{fontenc}

\usepackage{aecompl}
\usepackage{algorithm}
\usepackage[top=54pt, left=54pt, right=54pt, bottom=54pt]{geometry}
\usepackage{hyperref,url}
\usepackage{bm}
\usepackage{subfigure}
\usepackage{setspace}
\usepackage{booktabs}
\usepackage{cite}
\usepackage{amsmath,amssymb,amsfonts}
\usepackage{bbm}     
\usepackage{algorithm,algpseudocode}
\usepackage{float}
\usepackage{graphicx}
\usepackage{textcomp}
\usepackage{xcolor}
\usepackage{bbding}
\usepackage[capitalize]{cleveref}
\usepackage[numbers]{natbib}
\usepackage{multicol}
\usepackage{hyperref}
\usepackage{etoolbox}

\begin{document}
\IEEEoverridecommandlockouts

\title{\vspace*{14pt}Generalized Locomotion in Out-of-distribution Conditions with Robust Transformer
}

\author{Lingxiao Guo$^{1}$,\;\; Yue Gao$^{2\dag}$

\thanks{
 This work was supported by the National Natural Science Foundation of China (Grant No. 92248303 and No. 62373242), the Shanghai Municipal Science and Technology Major Project (Grant No. 2021SHZDZX0102), and the Fundamental Research Funds for the Central Universities.
 
$^{1}$Lingxiao Guo is with Shanghai Jiao Tong University, P.R. China, {\tt\small glx15534565855@sjtu.edu.cn}. }
\thanks{
$^{2}$Yue Gao is with MoE Key Lab of Artificial Intelligence and AI Institute, Shanghai Jiao Tong University, Shanghai, P.R. China, {\tt\small yuegao@sjtu.edu.cn}.}
\thanks{$\dag$ Corresponding author.}
}

\makeatletter

\let\@oldmaketitle\@maketitle%
\renewcommand{\@maketitle}{\@oldmaketitle%
    \centering
    \includegraphics[width=1.06\linewidth]{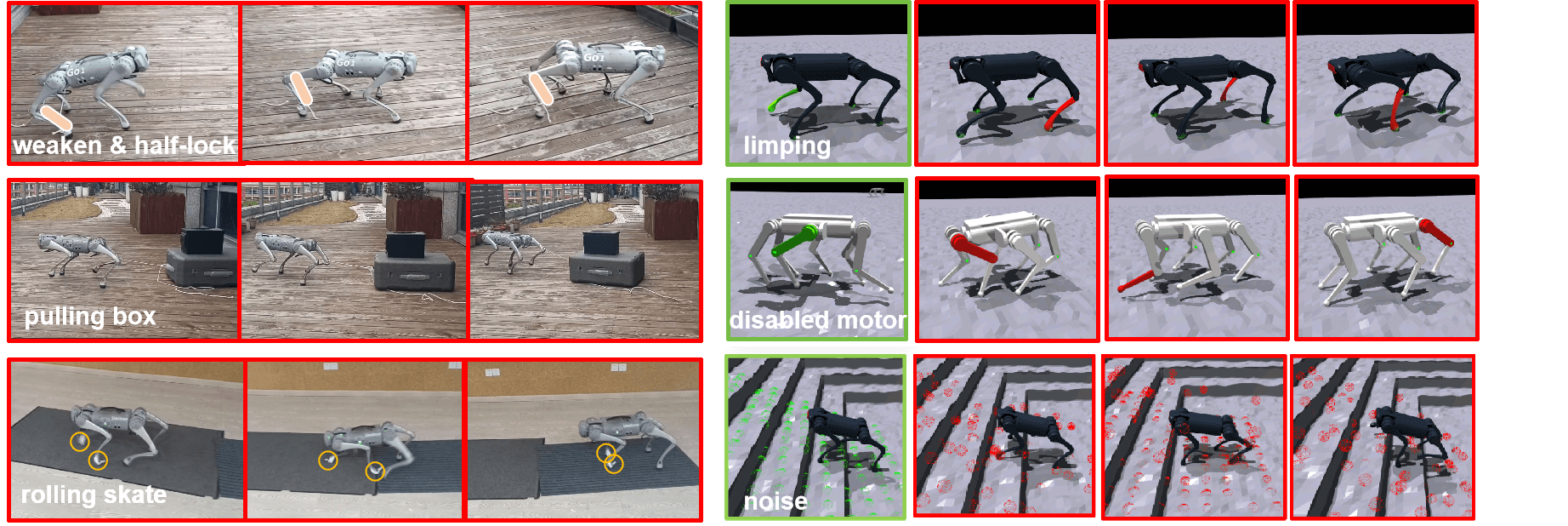}
    \raggedright
    \setstretch{0.75}
    \textbf{\footnotesize Fig 1. }{\footnotesize We present a locomotion controller to generalize under various out-of-distribution conditions. The pictures with green edges indicate the training distribution while the red ones indicate the test OOD distributions. While training under limited dynamic situations and clean observation, the controller can generalize to various unseen dynamics and sensor noise. }
    \label{fig:teaser}
    \vspace{-0.25cm}
}
\makeatother
\maketitle
\begin{abstract}
    To succeed in the real world, robots must deal with situations that differ from those seen during training. Those out-of-distribution situations for legged robot mainly include challenging  dynamic gaps and perceptual gaps. Here we study the problem of robust locomotion in such novel situations. While previous methods usually rely on designing elaborate training and adaptation techniques, we approach the problem from a network model perspective. Our approach, RObust Locomotion Transformer(ROLT),a variation of transformer,could achieve robustness in a variety of unseen conditions. ROLT introduces two key designs: body tokenization and consistent dropout. Body tokenization supports knowledge share across different limbs, which boosts generalization ability of the network. Meanwhile, a novel dropout strategy enhances the policy's robustness to unseen perceptual noise. We conduct extensive experiments both on quadruped and hexapod robots. Results demonstrate that ROLT is more robust than existing methods. Although trained in only a few dynamic settings, the learned policy generalizes well to multiple unseen dynamic conditions. Additionally, despite training with clean observations, the model handles challenging corruption noise during testing.
\end{abstract}

\section{Introduction}
A key challenge for robots in the real world is the distribution shift between training and deployment. Robots may encounter various out-of-distribution(OOD) situations that differ from training. For legged robots, the distribution shifts mainly involve two aspects: dynamic gap\cite{hwangbo2019learning} and perceptual gap\cite{miki2022learning}. While those gaps have been studied a lot by robot communities\cite{kumar2021rma}\cite{agarwal2023legged}\cite{lee2020learning}, solving tasks that may be OOD for all the training situations is rarely considered. Take the dynamics gap as a example. Consider a quadrupedal robot that has learned walking in a healthy condition during training and it's required to execute a search-and-rescue tasks in the real world. When traversing unstructured obstacles, the robot may stuck and damage its leg. Hence, the policy needs to be robust to such accidents and continue to complete the task. Similarly, sensor for perception may fail or produce noise. In contrast, the observation data in simulation is often accurate. Generally speaking, we expect to develop trustworthy robot controllers that can generalize and be robust to novel scenarios which are unseen during training.

Some methods use domain randomization during training to enhance the policy’s generalization for deployment. These methods train the policy across a range of dynamic parameters or under various observation noises\cite{kumar2021rma}\cite{agarwal2023legged}. Furthermore, a distilling training stage is often required to infer those dynamic parameters or observation ground truth. The robot can estimate dynamic and geometry information either through proprioception history\cite{lee2020learning}, noisy elevation maps\cite{miki2022learning} or depth images\cite{agarwal2023legged}. By inferring and adapting in real time, the policy can generalize to a range of dynamic and noisy conditions in unseen scenarios. There are also approaches that combine reinforcement learning with other methods, such as VAE or constrastive learning, to infer those parameters implicitly\cite{nahrendra2023dreamwaq}\cite{long2023hybrid}. Unlike these methods, we focus on solving tasks that are not covered by domain randomization. Since it's time-consuming to simulate all the situations that robot may encounter in the real world, it's important to develop a robust and generalized controller.

Recently, single life reinforcement learning\cite{chen2022you} is a promising paradigm for robots to adapt in the real world. This paradigm requires robot to adapt to new instances in a single episode. \citet{smith2022legged} finetuned the locomotion  policy in OOD conditions. By combining locomotion policy and a reset policy, the robot can conduct multiple trials in the real world. \citet{chen2023adapt} utilized prior behaviors to complete new tasks. They relied on a regularized value function to generalize to novel tasks. The robot dynamically select a suitable behavior module at each step. While all these approaches focus on designing training paradigms to develop a robust controlling procedure, we resort to network structure design to unleash its generalization ability. 

%最后面贡献那段前面总结的时候，加上下面这两个点
In this paper, we propose RObust Locomotion Transformer(ROLT), a model that is robust to unseen dynamic conditions and perceptual noise for locomotion tasks. Our insight is that the controller policy can have strong performance in novel scenarios by truly understanding the underlying nature of locomotion problem. Based on this view, we first design body tokenization for different limbs of the robot to share knowledge, and thus improve generalization. We then design a dropout strategy in different network layers, which facilitates a robust and flexible fusion between multi-modal perception. Our contributions are concluded as follows:

\begin{itemize}
\item We formulate ROLT, a robust multi-modal transformer. It takes prioproception and exteroception tokens as input, fusing them with attention mechanism.
\item Robot body tokenization and consistent dropout are created to fully unleash the transformer's generalization and robustness for locomotion.
\item Extensive experiments are conducted on quadrupedal and hexpodal robots. Results demonstrate that although trained only in a handful situations and observations with no noise, the policy could generalize to various unseen dynamic situations and perceptual noise.
\end{itemize}
\section{Related work}

{\textbf{Transformer for locomotion}}
Recently a series of works have applied transformer\cite{vaswani2017attention} for robot locomotion tasks. Given the strong sequence modeling capability of transformers, \citet{lai2023sim} and \citet{caluwaerts2023barkour} applied transformer to quadruped robots to infer motor actions from observation sequence. \citet{radosavovic2023learning} proposed a causal transformer for bipedal robots, which takes past observations and actions as input and predicts the next action. \citet{yang2021learning} leveraged a cross-modal transformer for robots to plan from depth images. While all these works introduced transformer to locomotion tasks, few of them considered more abilities of transformer, such as generalization and robustness. In contrast, transformer has shown great potential in generalization and robustness in computer vision\cite{naseer2021intriguing}\cite{dosovitskiy2020image}, natural language processing\cite{ontanon2021making} and reinforcement learning\cite{gupta2022metamorph}. We draw inspiration from those areas to design our robust controller.

\setcounter{figure}{1}
\begin{figure*}
	\centering
	\includegraphics[scale=0.33]{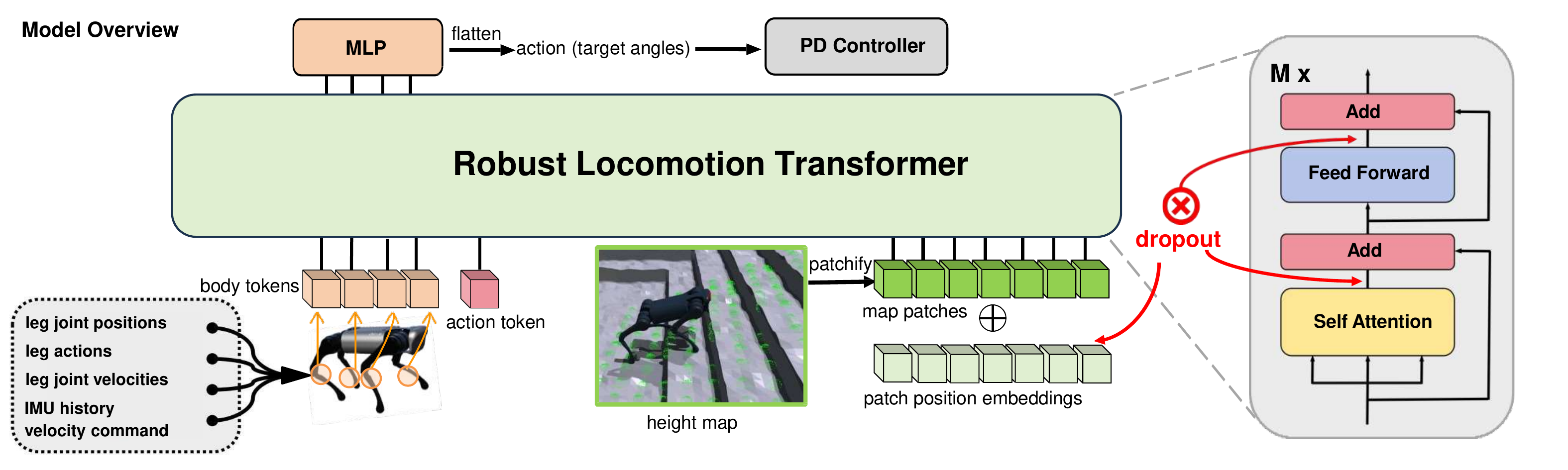}
 \vspace{-10pt}
	\caption{Model overview. We propose a multi-modal transformer to fuse exteroception and prioproception. The prioproception is tokenized into body tokens to support knowledge transfer across different limbs. An action token is fed into transformer to dynamically provide position information to the body tokens. Two consistent dropout operations, feature dropout and PE dropout to enhance the transformer's robustness.}
	\label{fig:encoder}
 \vspace{-15pt}
\end{figure*}

{\textbf{Dropout and robustness}}
Dropout\cite{srivastava2014dropout} is a technique to prevent network from overfitting. AI communities have proposed multiple dropout techniques to enhance the robustness of the model\cite{wu2021unidrop}\cite{shi2020informative}. While those approaches most pertain to supervised learning, consistent dropout\cite{hausknecht2022consistent} stabilizes dropout operations in reinforcement learning by applying the same dropout mask during rollout and training phase. However, it's still unclear about the role of dropout in reinforcement learning. On the other hand, \citet{campanaro2023roll} demonstrated dropout only during rollout phase could simulate sensor noise. Nevertheless, the dropout rate in \cite{campanaro2023roll} is extremely low, limiting its potential for adapting to robots. In this paper, we apply a simple yet efficient dropout strategy to the transformer. We demonstrate that our dropout method enables the policy to handle various challenging unseen noise during testing with minimal human intervention.

\section{Method}
We formulate the robot locomotion as a reinforcement learning framework. The state space is composed of two parts: proprioceptive state $s_p$ and external perception $s_e$. $s_p$ includes the robot’s body linear velocity, angular velocity, gravity projection and velocity command, as well as each joints' position, velocity and previous action. $s_e$ is the height values of scandots uniformly sampled around the robot. The action space is a target joint position. 

For the policy network, we propose a transformer architecture to dynamically fuse the external perception and proprioceptive state of legged robots. Figure 2 shows our main architecture. We first  modularize the body prioproception into tokens. Together with elevation map patches, they are processed by a lightweight transformer to fuse multi-modal perception. After that, updated body tokens are decoded through a linear layer to get the final actions. The following subsections will clarify the details of these methods.

\subsection{Body Tokenization}
To transform the robot’s state into a sequence, we tokenize the different modalities of the robot separately. First, similar to \cite{whitman2023learning}, we view the legged robot as a combination of limb modules. This leverages the fact that the morphology of a robot can be represented as a graph. Considering that the legs of legged robots have similar structures, we treat each leg as a token. This supports efficient dynamic and control knowledge sharing across different body tokens. For example, once a leg has a calf joint broken, and the robot learns to walk in such failure, one could expect the fault-tolerant experience mastered by this leg could be transferred to other legs. 

However, this generalization ability is hard to achieve by vanilla transformer.
Attention mechanisms often lack sensitivity to the position of tokens within a sequence. To counteract this, the transformer model requires the integration of Position Embeddings (PE). Commonly, as described in \cite{dosovitskiy2020image}\cite{gupta2022metamorph}, these embeddings are added directly to the tokens and are represented by data-independent, learnable vectors. However, in our specific settings, these traditional methods of position embedding significantly hinder the ability to transfer knowledge between different body limbs, because the leg tokens will overfit the data-independent PE. Those position embedding ways make different limb controller heterogeneous, and the experience mastered by one leg is hard to transfer to other legs.

To overcome the limitations of data-independent PE, we concatenate an action token to the leg tokens instead of adding learnable parameters directly to these tokens. This action token is a linear embedding of the previous step's action. The leg modules query their position information through the action token in the transformer. Importantly, this approach provides position information in a data-dependent manner. The leg tokens interact with the action tokens via attention, allowing them to infer their spatial positions flexibly. Consequently, this prevents the controller from overfitting the PEs, facilitates knowledge transfer between leg modules, and enhances the controller’s generalization.

In implementation, each leg token includes its past 10 steps of observation history. Then, a two layer MLP is utilized to project each token separately. Last target angles on all the joints are projected into the action token via a linear layer. The detailed tokenization schedule is formulated as follows:
\begin{align*}  e^{l_i} &= MLP(s^{l_i})  & e^{l_i}&\in\mathbb{R}{^D} \tag{1}\\ e^{a} &= W^{a}a_{t-1} + b^{a} & e^{a}&\in\mathbb{R}^{D}\tag{2}\\e^{prop} &= [e^{l_0}, e^{l_1}, …, e^{l_{L-1}},e^{a}, ] & e^{prop}&\in\mathbb{R}^{(L+1)\times D } \tag{3}\end{align*}
where $s^{l_i}$ refers to the observation history of the $i$th leg. It includes the past 10 steps of the joint positions, joint velocities and previous actions of all actuators on the $i$th leg, as well as the velocity command and base IMU observations. $a_{t-1}$ refers the last actions, i.e., the last output target angles. $e^{a}, e^{l_i}$ are embedding vectors corresponding to $a_{t-1}$ and $s^{l_i}$ respectively. $e^{prop}$ is proprioception token embedding, $D$ is the hidden dimension and $L$ is the number of legs.

\subsection{Transformer Encoder}
After modularizing the robot prioproception, we now clarify the patchify operations for the exteroception and the transformer controller structure. We use an elevation map as our external perception modality. Following the approach of ViT\cite{dosovitskiy2020image} for image tokenization, we reshape the elevation map $S^h \in \mathbb{R}^{H\times W}$ into a sequence of flattened 2D patches $s^h\in \mathbb{R}^{N^2\times \frac{H}{N} \times \frac{W}{N}}$, where $H$ and $W$ are the height and width of the elevation map, and $N^2$ is the number of patches. Then, we feed $s^h$ into a linear layer to obtain latents. To incorporate position information, learnable patch position embedding $p_e$ is added to those latents to obtain patch embedding $e^h \in \mathbb{R}^{N^2\times D}$. After getting all the tokens, we concatenate the proprioception token embedding $e^{prop}$ and the patch embedding $e^h$ to form the complete token embedding: $E_0 = [e^{prop},e^h]$, where $E_0\in \mathbb{R}^T$, and $T = L+N^2+1$ denotes the number of all tokens.

 Two transformer encoder layers are stacked to extract $E_0$ and output actions. The attention mechanism in transformer allows for flexible and robust fusion between tokens from different modalities. After updating token embeddings from the transformer encoder,  we directly use a linear layer to decode the tokens. Then we select the decoding results of the tokens corresponding to the leg modules to obtain the joint target angles $a_t\in\mathbb{R}^{L\times J}$, where J is the number of joints on each leg. Then we flatten $a_t$ and feed it to the corresponding joint PD controller. 

\subsection{Consistent Dropout}
While it's hard to simulate the challenging noise during test, we design a novel dropout strategy to stimulate the model's robustness against multiple noise. The dropout strategy is conducted in a consistent way following\cite{hausknecht2022consistent}. For the traditional dropout technique and Roll-drop\cite{campanaro2023roll}, the use of different dropout masks for the same state at rollout time versus update time can lead to vastly different action probabilities and potential training failures\cite[Section 4]{campanaro2023roll}. As the dropout rate increases, the difference of action probabilities becomes more pronounced, destabilizing policy training. We mitigate this problem by applying the same dropout mask during rollout phase and update phase. This allow us to adopt a much larger dropout rate in reinforcement learning. So we could better unleash the potential of dropout for robots. Specifically, our dropout strategy consists of two operations: feature dropout and PE dropout.

{\textbf{Feature dropout}}
Feature dropout refers to randomly suppress neurons of neural networks during training by setting them to 0 with a dropout rate $p$. Through this dropout operation, randomness is injected to the policy. During training, the policy must be robust enough to utilize multiple subset of neurons to make proper decisions. Note that the randomness injected by feature dropout is domain-agnostic. This improves the controller's robustness to  various unseen noise with minimal human knowledge. Specifically, we drop features both after self-attention and feed-forward network.

{\textbf{PE dropout}}
Besides feature dropout, we also dropout patch position embedding $p_e$ in a high rate. This means we use a single learnable vector to replace all dropout position embeddings during training. And during test, we apply position embeddings to all the patches. This is because the exteroception may contain various noises which deviate significantly from the true value. Directly using the spatial position information offered by position embedding may lead to misleading results. On the other hand, the robot’s proprioception is usually more precise than exteroception. The robot can infer the spatial information of the patches through the interaction of its proprioception and exteroception in self-attention. This enables the robot to estimate the terrain geometries in a more reliable way. 
\begin{table}[H]
    \centering
    \small
    \caption{Reward terms}
    \label{tbl:rewards}
    \vspace{-5pt}
    \begin{tabular}{llr}
        \toprule
        Term & Symbol & Equation \\ [0.5ex]
        \midrule
        xy velocity tracking  & \( \exp\left(-\frac{\textbf{v}_{x,y} - \textbf{v}^{\text{cmd}}_{x,y}}{\sigma_v} \right) \) & 1.5\\ [0.5ex]
        yaw velocity tracking & \( \exp\left(-\frac{\boldsymbol{\omega}_z - \boldsymbol{\omega}^{\text{cmd}}_{z}}{\sigma_{\omega z}} \right) \) & 0.5 \\ [0.8ex]
        z velocity & \( \textbf{v}_{z}^2 \) & -2 \\ [0.5ex]
        roll-pitch velocity & \( |\boldsymbol{\omega}_{xy}|^2 \) & -0.05 \\ [0.5ex]
        collision & \( \mathbbm{1}_{\text{collision}} \) & -1 \\ [0.5ex]
        joint torques & \( |\boldsymbol{\tau}|^2 \) & -2\text{e}{-4} \\ [0.5ex]
        action rate & \( |\textbf{a}_{t} - \textbf{a}_{t-1}|^2 \) & -0.01 \\ [0.5ex]
        foot airtime & \( \sum t_{\text{air}} \cdot \mathbbm{1}_{\text{new contact}} \) & 1 \\ [0.5ex]
        joint error & \( |\textbf{q}_{t} - \textbf{q}_{\text{default}}|^2 \) & -0.04 \\ [0.5ex]
        \bottomrule
    \end{tabular}
\end{table}
\vspace{-10pt}
{\textbf{Training details}}
To reduce the computational cost on the hardware, we make the proposed transformer as lightweight as possible. Our transformer encoder consists of 2 encoder layers, each of which has a hidden dimension of 160. The feature dropout rate is $0.1$ and the PE dropout rate is $0.75$. We train our network using the standard PPO\cite{schulman2017proximal} algorithm. The reward functions are listed in Table \ref{tbl:rewards}.

\begin{figure*}
\vspace{-10pt}
	\centering
	\subfigure[Stiffness test: Distance]{			 
   \includegraphics[scale=0.35]{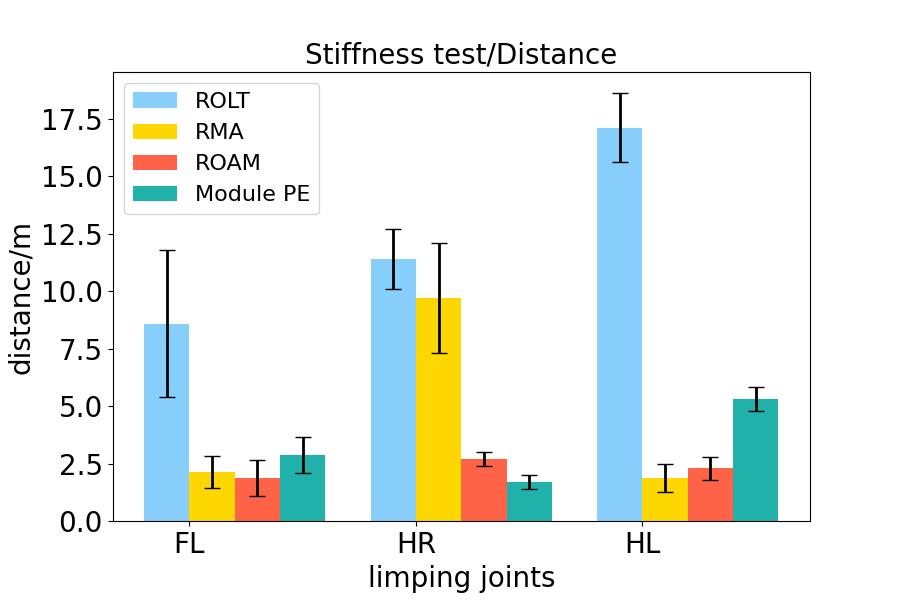}\label{a1stiffness_distance}}		
\subfigure[Stiffness test: Fall time]{		
  \includegraphics[scale=0.35]{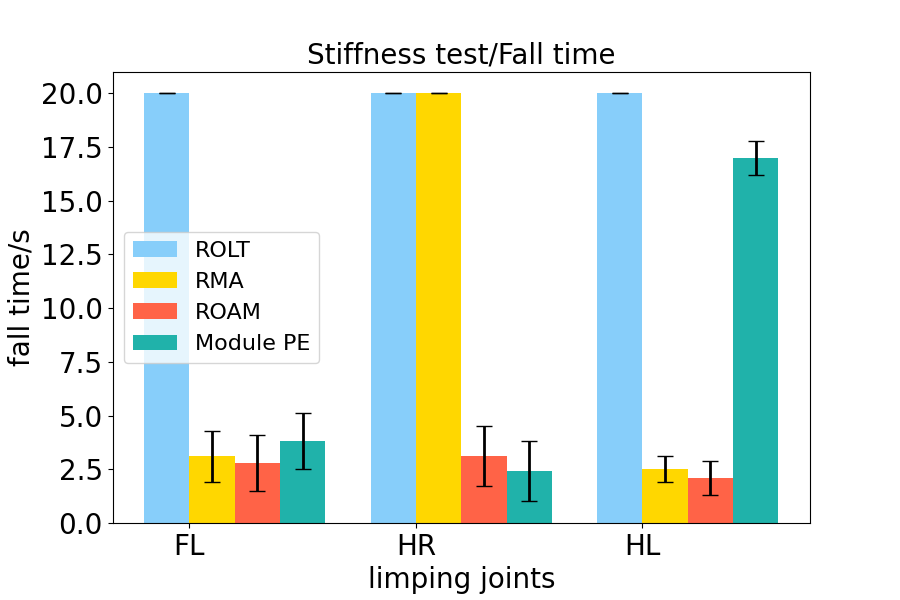}\label{a1stiffness_fall_time}}	
  \vspace{-10pt}
\caption{Stiffness test results}
\vspace{-10pt}
\label{Stiffness test results}
\end{figure*}
 \vspace{-20pt}
\begin{table*}[t]
\caption{{Damage test results on the hexapodal robot }} 
\label{Damage_test}
\centering
\resizebox{\textwidth}{!}{
\begin{tabular}{lccccccccc}
\toprule
Damage joints& FLC(calf)& FLT(thigh)& FRC& MRC & MLT& HLT& HLC& HRT & HRC\\\midrule
 Method & \multicolumn{9}{c}{Distance /m$\uparrow$ (distance that the robot walked in 20s given a velocity command at 0.75m/s)  }\\
\midrule
RMA& $1.83 \scriptstyle{\pm 0.77 }$ & $2.56 \scriptstyle{\pm 2.11 }$ & $5.52 \scriptstyle{\pm 2.95 }$ & $13.36 \scriptstyle{\pm 3.25 }$ & $13.83 \scriptstyle{\pm 2.31 }$ & $2.36 \scriptstyle{\pm 1.53 }$ & $4.41 \scriptstyle{\pm 2.39 }$ & $3.95 \scriptstyle{\pm 1.63 }$ & $4.33 \scriptstyle{\pm 2.04 }$ \\
ROAM& $1.78 \scriptstyle{\pm 1.41 }$ & $2.26 \scriptstyle{\pm 1.21 }$ & $15.18 \scriptstyle{\pm 3.17 }$ & $14.19 \scriptstyle{\pm 2.81 }$ & $13.91 \scriptstyle{\pm 3.28 }$ & $8.16 \scriptstyle{\pm 3.19 }$ & $4.29 \scriptstyle{\pm 2.17 }$ & $4.18 \scriptstyle{\pm 2.47 }$& $3.28 \scriptstyle{\pm 2.19 }$ \\
Module PE& $\mathbf{4.28} \scriptstyle{\pm 2.93 }$ & $ \mathbf{3.39} \scriptstyle{\pm 1.26 }$ & $\mathbf{14.38} \scriptstyle{\pm 2.78 }$ & $\mathbf{15.06} \scriptstyle{\pm 2.90 }$ & $\mathbf{11.75} \scriptstyle{\pm 2.84 }$ & $\mathbf{14.60} \scriptstyle{\pm 2.76 }$ & $\mathbf{1.90} \scriptstyle{\pm 0.84 }$ & $\mathbf{13.32} \scriptstyle{\pm 3.50 }$ & $\mathbf{3.89} \scriptstyle{\pm 2.59 }$ \\
ROLT(Ours)& $\mathbf{14.28} \scriptstyle{\pm 2.93 }$ & $ \mathbf{13.39} \scriptstyle{\pm 2.26 }$ & $\mathbf{14.68} \scriptstyle{\pm 3.78 }$ & $\mathbf{16.06} \scriptstyle{\pm 2.10 }$ & $\mathbf{15.15} \scriptstyle{\pm 2.84 }$ & $\mathbf{12.56} \scriptstyle{\pm 3.26 }$ & $\mathbf{12.60} \scriptstyle{\pm 2.74 }$ & $\mathbf{15.32} \scriptstyle{\pm 1.50 }$ & $\mathbf{13.89} \scriptstyle{\pm 2.64 }$ \\
\midrule
& \multicolumn{9}{c}{Fall time/s $\uparrow$ }\\
\midrule
RMA& $2.31 \scriptstyle{\pm 0.89 }$ & $3.56 \scriptstyle{\pm 2.81 }$ & $7.52 \scriptstyle{\pm 4.35 }$ & $20.0 \scriptstyle{\pm 0.0 }$ & $20.0 \scriptstyle{\pm 0.0 }$ & $3.45 \scriptstyle{\pm 2.13 }$ & $4.64 \scriptstyle{\pm 2.89 }$ & $5.15 \scriptstyle{\pm 2.67 }$ & $6.74 \scriptstyle{\pm 3. }$ \\
ROAM& $2.26 \scriptstyle{\pm 1.31 }$ & $3.17 \scriptstyle{\pm 0.81 }$ & $20.0 \scriptstyle{\pm 0.0 }$ & $20.0 \scriptstyle{\pm 0.0 }$ & $20.0 \scriptstyle{\pm 0.0 }$ & $11.16 \scriptstyle{\pm 4.28 }$ & $6.29 \scriptstyle{\pm 3.17 }$ & $4.18 \scriptstyle{\pm 3.47 }$& $2.28 \scriptstyle{\pm 11.19 }$ \\
Module PE& $\mathbf{5.98} \scriptstyle{\pm 3.53 }$ & $ \mathbf{4.51} \scriptstyle{\pm 1.67 }$ & $\mathbf{20.0} \scriptstyle{\pm 0.0 }$ & $\mathbf{20.0} \scriptstyle{\pm 0.0 }$ & $\mathbf{20.0} \scriptstyle{\pm 0.0 }$ & $\mathbf{20.0} \scriptstyle{\pm 0.0 }$ & $\mathbf{2.87} \scriptstyle{\pm 2.31 }$ & $\mathbf{20.0} \scriptstyle{\pm 0.0 }$ & $\mathbf{4.49} \scriptstyle{\pm 2.34 }$ \\
ROLT(Ours)& $\mathbf{20.0} \scriptstyle{\pm 0.0 }$ & $ \mathbf{20.0} \scriptstyle{\pm 0.0 }$ & $\mathbf{20.0} \scriptstyle{\pm 0.0 }$ & $\mathbf{20.0} \scriptstyle{\pm 0.0 }$ & $\mathbf{20.0} \scriptstyle{\pm 0.0 }$ & $\mathbf{20.0} \scriptstyle{\pm 0.0 }$ & $\mathbf{20.0} \scriptstyle{\pm 0.0 }$ & $\mathbf{20.0} \scriptstyle{\pm 0.0 }$ & $\mathbf{20.0} \scriptstyle{\pm 0.0 }$ \\
\bottomrule

\end{tabular}
}

\end{table*}

\vspace{20pt}
\section{Experiments}
\textbf{Experiment setup: }We train 4096 parallel agents in the IsaacGym simulator with the legged gym library\cite{rudin2022learning}. The policy
 is trained with 400 million
 simulated time steps. We evaluate the generality and robustness of our approach by conducting simulation experiments on Unitree A1 and a hexapodal robot. The overall training time for A1 and the hexapodal robot is about four hours and six hours of wall-clock time. All trainings are performed on a single NVIDIA RTX 3090 GPU. We conduct real-world experiments on Unitree Go1. We design experiments to answer following questions:

 \begin{enumerate}
 \item Can ROLT generalize to various OOD dynamic conditions while training only in limited situations?
 \item Can ROLT be robust to various perceptual noise while training only under clean observations?
 \item What patterns does ROLT learn about the locomotion? 
 \item Can ROLT maintain robustness in the real world?
\end{enumerate}

\textbf{Baseline and ablations: }We compare ROLT with several baselines and ablations. These baselines include \textbf{MLP}, \textbf{Roll-Drop}\cite{campanaro2023roll}, \textbf{RMA}\cite{kumar2021rma} and \textbf{ROAM}\cite{chen2023adapt}. These ablations include \textbf{Module PE}, \textbf{No PE Drop} and \textbf{No Feature Drop}.
\begin{itemize}
\item \textbf{MLP: } it trains a naive MLP with PPO. The MLP takes both prioprocetion and exteroception as input. 
\item \textbf{Roll-Drop: } the network is the same to \textbf{MLP}, but the neurons are dropout during rollout in a rate of $0.0001$.
\item \textbf{RMA: } it contains an adaptation module to infer dynamic parameters and a base policy to output actions. We additionally adopt a CNN to encode height values in our setting. Gussian noise are added during training.
\item \textbf{ROAM: } it utilizes a regularized value function to select a suitable behavior module in novel situations. The behavior modules are acquired through training MLP in corresponding situations.
\item \textbf{Module PE: }it injects the position information to the limb modules by adding a set of learnable parameters to the tokens instead of incorporating an action token.
\item \textbf{No PE Drop: }ROLT without PE dropout operations.
\item \textbf{No Feature Drop: } ROLT without the feature dropout.
\end{itemize}
\begin{figure*}
\vspace{-1pt}
	\centering
	\subfigure[Weaken\&half-lock: ROLT]{			 
   \includegraphics[scale=0.205]{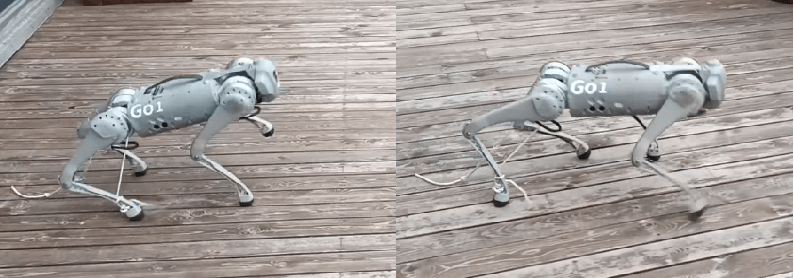}\label{real-rolt-weaken}}	
 \subfigure[Pulling box: ROLT]{			 
   \includegraphics[scale=0.20]{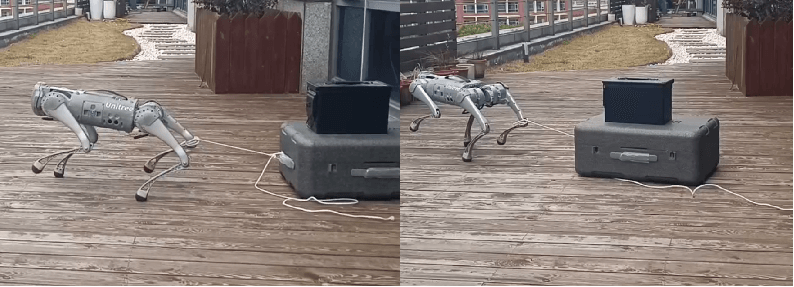}\label{real-rolt-weaken}}	
 \subfigure[Rolling skate: ROLT]{			 
   \includegraphics[scale=0.32]{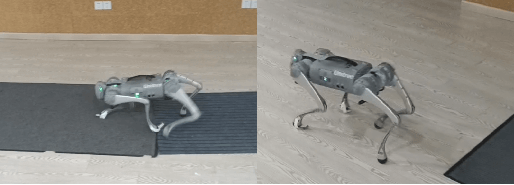}\label{real-rolt-roller}}	
\subfigure[Weaken\&half-lock: RMA]{		
  \includegraphics[scale=0.42]{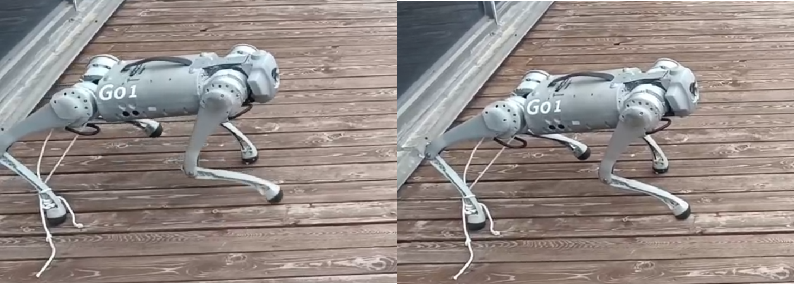}\label{real-rma-weaken}}			
\subfigure[Pulling box: RMA]{		
  \includegraphics[scale=0.2]{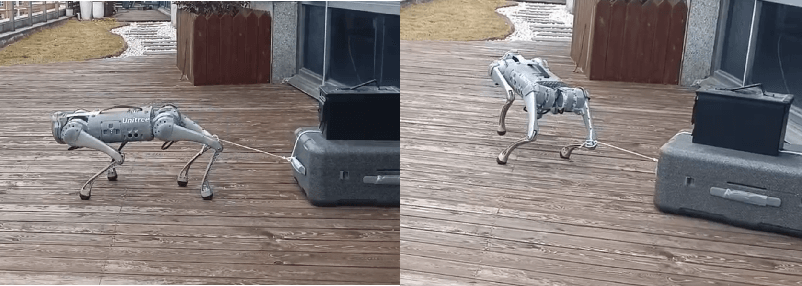}\label{real-rma-weaken}}	
\subfigure[Rolling skate: RMA]{		
  \includegraphics[scale=0.32]{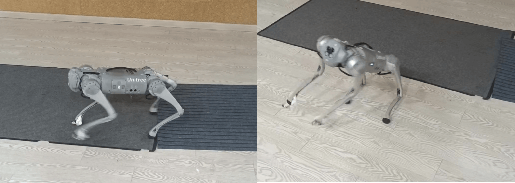}\label{real-rma-weaken}}	
  \vspace{-15pt}
\caption{Real World Experiments}
\vspace{-15pt}
\label{Real World Experiments}
\end{figure*}

\subsection{\textbf{Generalization in OOD dynamics}}
To answer (1), we design two experiments on A1 and the hexapodal robot: (i) Stiffness test on A1, where the front right leg limps during training and the other legs limp in the test. To simulate limping we simply lock the calf joint on corresponding legs. (ii) damage test on the hexapodal robot, where in training, we randomly disable one joint motor among the thigh joint on front right leg(FRT), the thigh joint on mid right leg(MRT), and the calf joint on mid left joint(MLC). For disabling, we simply set the torque output to zero. In test, we disable one joint from all the rest calf and thigh joints. All the training process additionally include a normal healthy walking situation. For evaluation, we give a velocity command at 0.75m/s, test the distance that the robot walked in 20s and record time of robot's first fall.
\begin{figure*}
 \vspace{-2pt}
	\centering
	\includegraphics[scale=0.27]{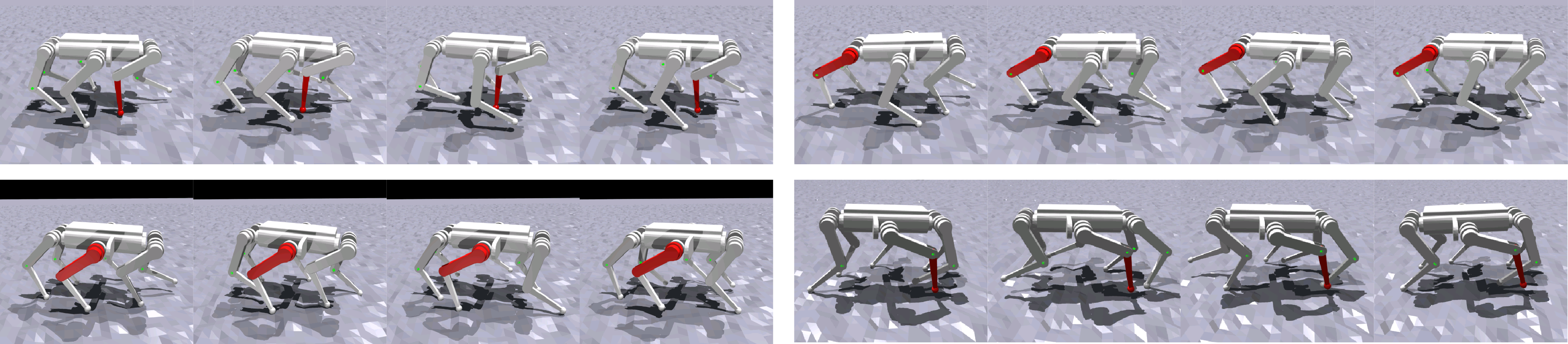}
  \vspace{-5pt}
	\caption{Unstructured gaits shown by ROLT when the hexapodal robot faces unseen motor damages.}
	\label{fig:encoder}
 \vspace{-20pt}
 \label{gaits}
\end{figure*}

\textbf{Stiffness test: }The results of stiffness test are shown in Figure \ref{Stiffness test results}. ROLT has demonstrated surprising generalization ability across different dynamics than other methods. By training only in healthy conditions and FR limping situations, ROLT could still maintain walking in a fault-tolerant way when two new legs are limping. This proves that through proper design, even robot controllers can have the ability to raise one and infer three. In contrast, RMA quickly falls in two out of three failures. This is mainly because their networks can't capture the underlying problem structure in this settings. Locking one leg can't borrow experience for other legs, so they fail to generalize to other limping situations. We also note that ROAM fails to walk in all three test failures. This is mainly because each behavior module of ROAM is trained independently within its respective scenario. So the generalization of each module is limited. While ROAM relies on choosing most familiar behaviors according to the value function, it fails to query a suitable policy due to the too few behavior modules. For the ablations, we note that the Module PE performs nearly as awful as RMA. This is because that the transformer is inclined to overfit the leg PE. Consequently, each leg policy receives a distinctly different signal, which prevent different limb modules for knowledge sharing. This demonstrates dynamically querying position information in action token is essential for generalization.
% 可能要补一下步态分析.肯定要补了。。。

%这一段应该也要补步态分析
\textbf{Damage test: }We show damage test results in Table \ref{Damage_test}. By training in a handful of damaged situations, ROLT is competent to generalize to various failures in the leg joints. Note that the hexapodal robot has redundant joints, and hence has a potential for fault-tolerant working. ROLT could unleash this potential. Furthermore, this could be very useful when legged robots are required to execute risky missions, where unseen failures may occurs on the robot. On the contary, all the other methods show limited generalization.

To better understand the generalization brought by ROLT, we visualize the emergent gaits under different OOD situations in Figure \ref{gaits}. We observe that the ROLT learns unstructured gaits across different failures. The gaits support multiple beats to the ground in a stride. When new damages occurs on legs, the robot learns to adapt the phase and frequency of each leg to walk safely. Although in most instances we don’t observe a regular gait in the hexapodal, this walking pattern shows robustness across different failures. On the other hand, we observe other methods often converge to a tripod gait in healthy situations. But when new damage occurs, the robot often exhibits wired gait such as stretching the leg and quickly fall. This demonstrates that ROLT has a better understanding of the leg motions. 

%比较MLP, RMA, Roll-Drop, No PE Drop
\subsection{\textbf{Robustness to unseen noise}}
To answer (2), we train A1 on multiple terrains, including stairs, slopes and platforms, with clean elevation observation. The map is split into 12 patches. Tests are conducted in stairs, with a fixed velocity command at $0.5m/s$. We inject two kinds of noise to the elevation map in test: (i).low-frequency noise, where a certain number of patches are offset as a whole by a distance in random directions; (ii).high-frequency noise, where every height point is shifted vertically and laterally with Gaussian noise independently. The reward returns and distance walked in an episode are used for evaluation.

\begin{figure}
\vspace{0pt}
	\subfigure{			 
   \includegraphics[scale=0.23]{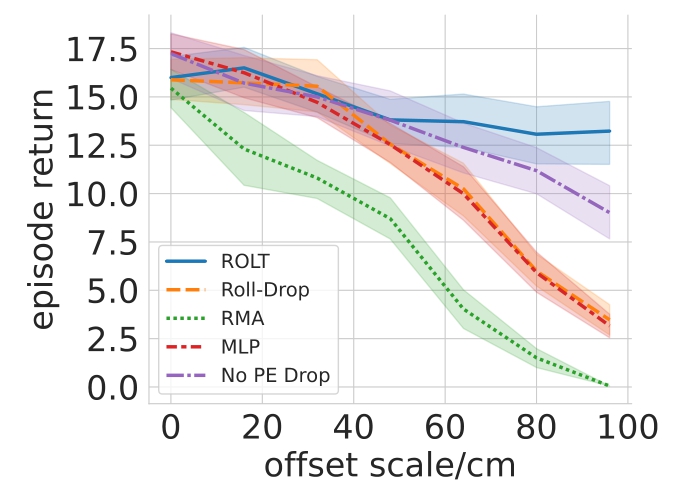}\label{a1OffsetReturn}}		
\subfigure{		
  \includegraphics[scale=0.22]{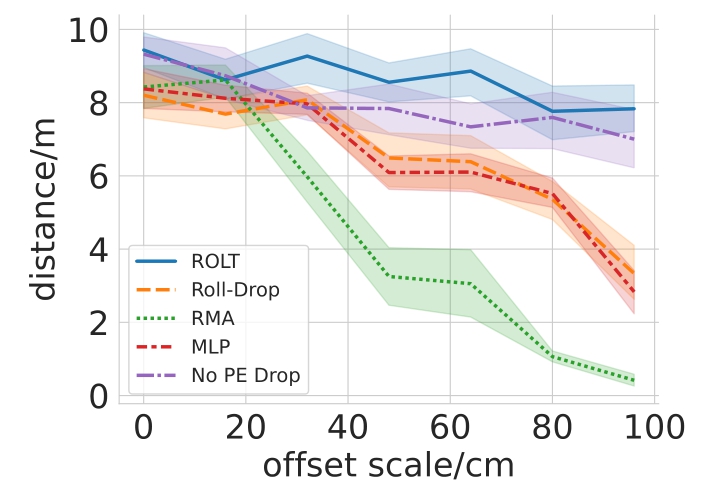}\label{a1OffsetDistance}}	
  \vspace{-10pt}
\caption{High-frequency noise test results} 
\label{Abnormal values}
\vspace{-10pt}
\end{figure}

\textbf{Low-frequency noise: }The results are shown in Figure \ref{Abnormal values}. As the offset scale increase, both ROLT maintain the best performance. Note that while both ROLT and Roll-Drop adopt dropout to inject randomness into the network, the performance of Roll-Drop decreases significantly when the offset scale increases. This is because the dropout strategy cannot resist low-frequency noise. However, attention mechanism is flexible to handle the low-frequency noise. When some height patches are abnormal, the attention could neglect those information by querying trustworthy proprioception. 

\begin{figure}
\vspace{-3pt}
	\subfigure{			 
   \includegraphics[scale=0.23]{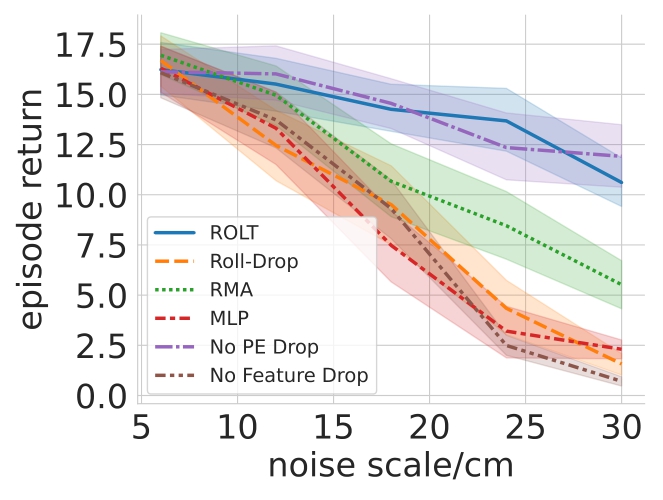}\label{a1NoiseReturn}}		
\subfigure{		
  \includegraphics[scale=0.23]{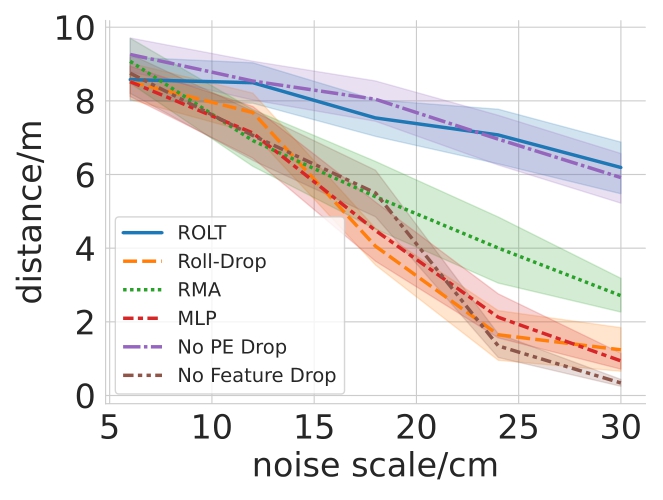}\label{a1NoiseDistance}}
   \vspace{-12pt}
\caption{Low-frequency noise test results}
\label{Gaussian noise}
 \vspace{0pt}
\end{figure}

\textbf{High-frequency noise: }The results are shown in Figure \ref{Gaussian noise} As the noise scale increase, ROLT achieves better performance than other approaches. Even MLP and RMA are injected with Gaussian noise during training, they still fail to generalize higher Gaussian noise. This indicates our dropout strategy is effective in injecting randomness and enhance robustness. Roll-Drop performs badly since it adopts a extremely small dropout rate and injects only limited randomness to the policy. The No Feature Drop ablations performs the worst, which proves that the feature dropout strategy is important to resist high-frequency noise.

\begin{table}[H]
\vspace{-7pt}
\caption{Real-world Experiments}
\label{Real world results}
\vspace{-7pt}
\centering
% \resizebox{\textwidth}{!}{
\begin{tabular}{lcccc}
\toprule
\multicolumn{4}{c}{\quad \quad \quad \quad \quad \quad Distance/m  $\uparrow$ }\\
\midrule
Method &Weaken \& half-lock& Pulling box& Rolling skate \\\midrule
RMA& ${0.0 \scriptstyle{\pm 0.0 }}$ & ${0.4 \scriptstyle{\pm 0.1}}$ & ${1.2 \scriptstyle{\pm 0.1 }}$ \\
ROLT&$\mathbf{2.8 \scriptstyle{\pm 0.2 }}$ & $\mathbf{1.5 \scriptstyle{\pm 0.3 }}$ & $\mathbf{4.8 \scriptstyle{\pm 0.2 }}$   \\

\bottomrule
\vspace{-15pt}
\end{tabular}

\vspace{-0.1in}
\end{table}
\vspace{-5pt}
\begin{figure}
	\centering
 \vspace{-5pt}
	\subfigure[]
  {			 
   \includegraphics[scale=0.2]{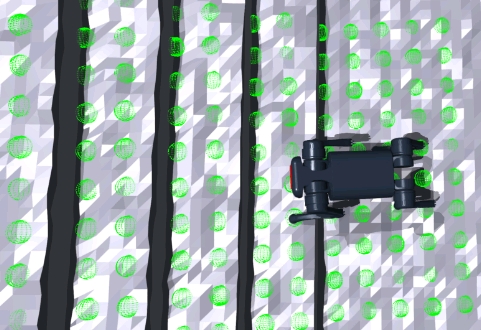}\label{kxy}}	
   \vspace{-5pt}
\subfigure[]{			 
   \includegraphics[scale=0.1]{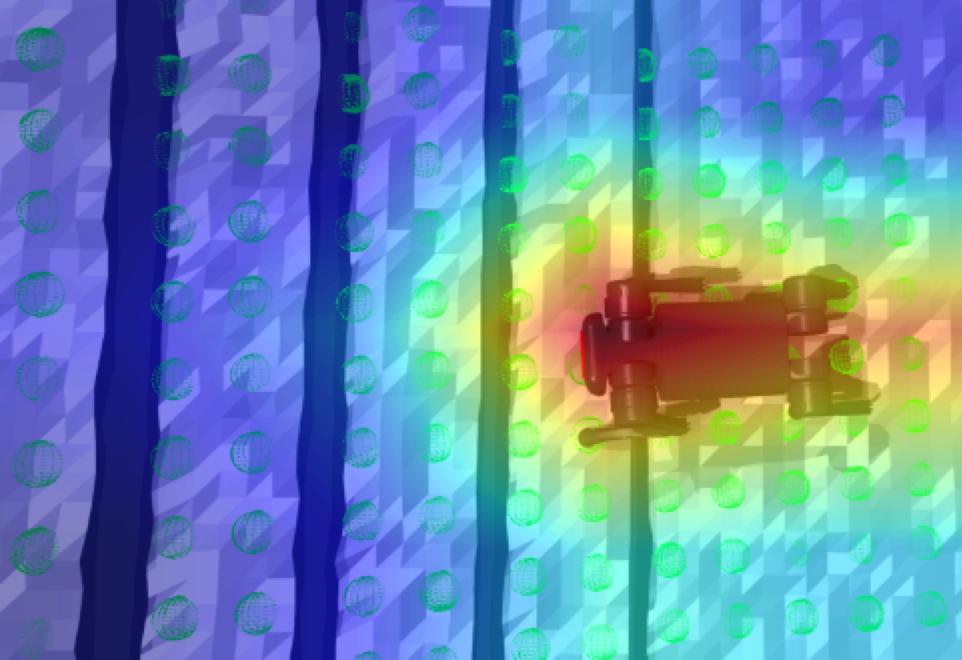}\label{kxy}}		
   \vspace{-5pt}
\subfigure[]{			 
   \includegraphics[scale=0.1]{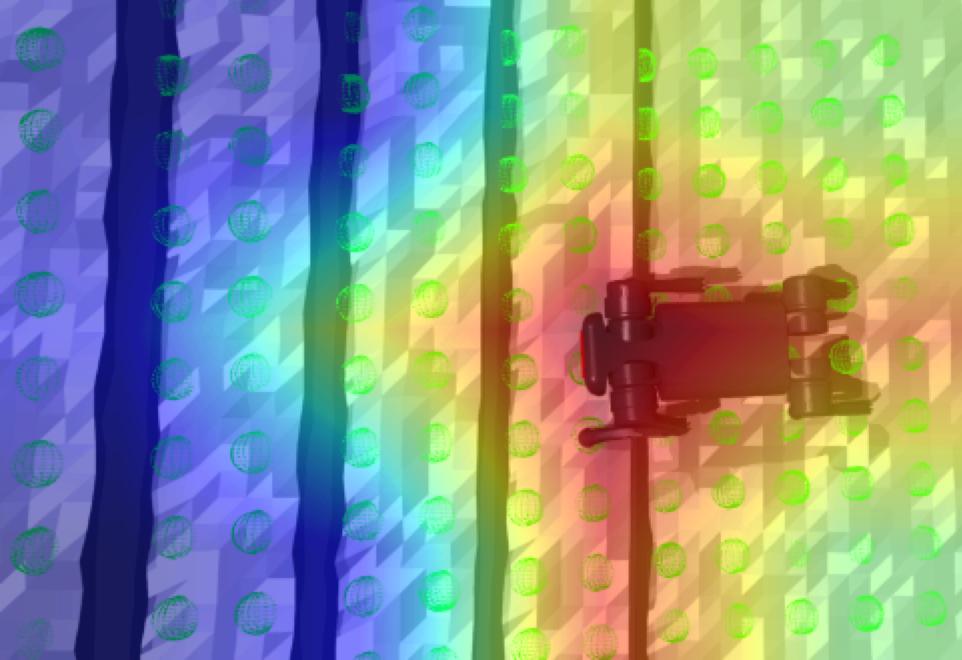}\label{kxy}}		
   \vspace{-5pt}
\subfigure[]{			 
   \includegraphics[scale=0.24]{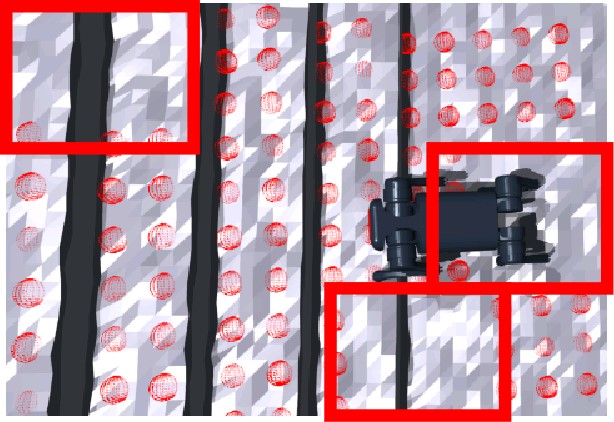}\label{kxy}}		
\subfigure[]{			 
   \includegraphics[scale=0.1]{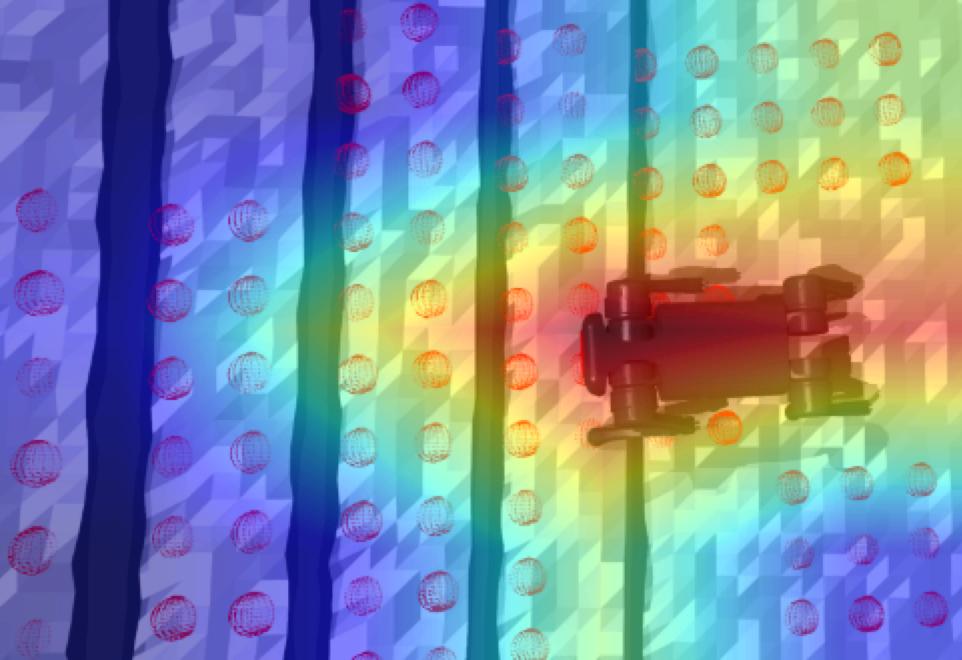}\label{kxy}}
\subfigure[]{			 
   \includegraphics[scale=0.1]{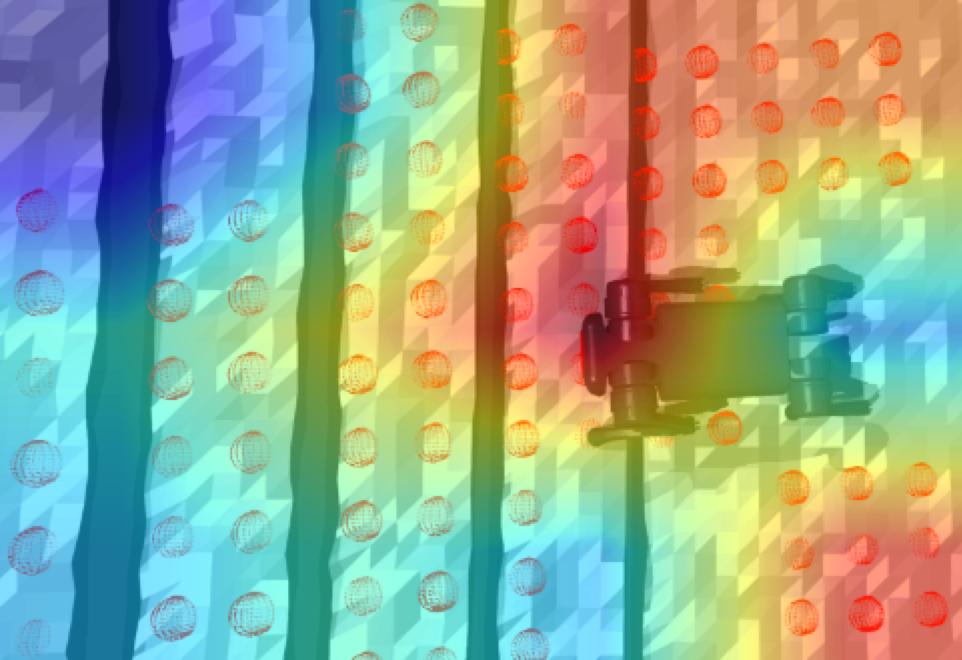}\label{kxy}}	
\vspace{-5pt}
	\caption{Elevation maps and attention maps in different situations. \textbf{(a)} clean observations. \textbf{(b)} attention map of No PE drop on (a). \textbf{(c)} attention map of ROLT on (a). \textbf{(d)} observations with missing patches \textbf{(e)} attention map of No PE Drop on (d).  \textbf{(f)} attention map of ROLT on (d).}
	\label{attention map}
\vspace{-20pt}
\end{figure}
% 可能要补一个prioproception实验，如果图的篇幅不够,也可以仅仅是多加点评价指标
   
\subsection{\textbf{Real world experiments}}
\vspace{-2pt}
We train our model with only the front right leg being limping on Go1 and zero-shot transfer the learned policy to physical Unitree Go1. Then we conduct three novel experiments to test its generalization under OOD conditions: 1. weaken\&lock: we weaken the motor strength of one leg and lock the hind back leg with a string, so the calf joint angle can't be larger than the locked angle; 2. pulling box: the robot has to pull a box with one single hind right leg; 3. rolling skates: two roller skates are tied to the robot's front feet. We compare it with RMA. For evaluation we count the distance that the robot walks in 10 seconds.

The results are shown in Table \ref{Real world results}.Our approach outperforms RMA. As Figure \ref{Real World Experiments} shown, in weaken\&lock settings, ROLT initially faltered due to dynamic changes but quickly adapted and resumed normal walking after a few steps. This proves that ROLT could effectively utilize the experience under one leg limping to handle new leg failures. In contrast, RMA failed to move, which shows RMA is bad at leveraging such knowledge and generalize. Thus it got stuck in the hind back leg and failed to utilize the locked leg to walk. In pulling box settings, it's hard to drag the heavy box with one leg since the box apply a large lateral force on the tied leg, causing an angular momentum applied to the robot's body. But ROLT is competent to maintain its balance and walk a distance, while RMA couldn't move straight forward and spin around the box.  For the rolling skate test, ROLT could recover from sudden sliding and steadily move forward. RMA slid laterally and finally fell. It turns out that ROLT could generalize to novel situations in the real world deployment using limited failure experience in the simulation, while others fail. 
\vspace{-3pt}
\subsection{\textbf{Visualization Analysis}}
\vspace{-2pt}
\label{Visualization Analysis}
 Visualization of the attention for the elevation map under different conditions are analyzed. Note that the attention contains tokens from multiple modalities, the attention score of the map patches on each leg token is averaged. Results are shown in Figure \ref{attention map}. Under clean observations, the No PE drop focus attention on the terrain surrounding its body. When the patches are masked to zero, it still pays attention to the similar spatial position in the map. This shows that the self-attention overfits the patch PE and cannot shift attention when noise is injected. In comparison, ROLT pays attention to more patches that is similar to the terrain near its body in normal settings. When some patches are masked, it will pay more attention to those unmasked to estimate terrain information. This implies that the PE dropout allows for a more flexible flow of attention through the surrounding terrain.

\vspace{-5pt}
\section{Conclusion} 
\vspace{-5pt}
\label{sec:conclusion}
    We present Robust Locomotion Transformer in this paper. The transformer employs body tokenization to achieve knowledge transfer across limbs and utilizes consistent dropout to enhance robustness. Extensive experiments show that our model exhibits strong generalization and robustness in novel dynamics and in unseen noise although only trained in limited situations. We believe our approach provides insight into designing network structures to achieve general robustness in controllers. For future directions, it's exciting to explore more generalization abilities of the controller, such as traversing new obstacles and generalizing novel motor skills.

%% Use plainnat to work nicely with natbib. 
\begin{footnotesize}
\bibliographystyle{IEEEtranN}
\bibliography{references}
\end{footnotesize}

\end{document}